\documentclass[11pt]{article}

\usepackage{nodalida2025}
\aclfinalcopy

\usepackage{hyperref}
\usepackage{url}
\usepackage{times}
\usepackage{latexsym}
\usepackage{graphicx}
\usepackage{multirow}
\usepackage{textcomp}  
\usepackage{scalerel}  

\def\check{\scalerel*{\includegraphics{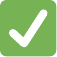}}{\textrm{\textbigcircle}}}
\def\cross{\scalerel*{\includegraphics{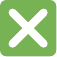}}{\textrm{\textbigcircle}}}
\def\painting{\scalerel*{\includegraphics{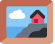}}{\textrm{\textbigcircle}}}
\def\paper{\scalerel*{\includegraphics{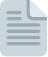}}{\textrm{\textbigcircle}}}

\usepackage[T1]{fontenc}

\usepackage[utf8]{inputenc}

\usepackage{microtype}

\usepackage{inconsolata}

\title{Image-Text Relation Prediction for Multilingual Tweets}

\author{
    Matīss Rikters, Edison Marrese-Taylor \\
    Artificial Intelligence Research Center, \\
    National Institute of Advanced Industrial Science and Technology \\
    \texttt{\{firstname.lastname\}@aist.go.jp} \\
}

\begin{document}
\maketitle
\begin{abstract}
    Various social networks have been allowing media uploads for over a decade now. Still, it has not always been clear what is their relation with the posted text or even if there is any at all. In this work, we explore how multilingual vision-language models tackle the task of image-text relation prediction in different languages, and construct a dedicated balanced benchmark data set from Twitter posts in Latvian along with their manual translations into English. We compare our results to previous work and show that the more recently released vision-language model checkpoints are becoming increasingly capable at this task, but there is still much room for further improvement. 
\end{abstract}

\section{Introduction}

Twitter (now X\footnote{From Twitter to X: Elon Musk Begins Erasing an Iconic Internet Brand - \url{https://www.nytimes.com/2023/07/24/technology/twitter-x-elon-musk.html}}) remains a crucial platform in modern society due to its role in shaping public discourse, enabling real-time communication, and fostering global conversations. As a microblogging site, it allows individuals, organizations, and governments to share thoughts, news, and opinions instantaneously. Even though potential alternatives have recently risen in popularity, they still exhibit distinct drawbacks to the general public, like Threads refusing to promote real-time content and news events, or Mastodon being too granulated and slow overall due to being dependent on the performance of individual servers. 

The integration of images with tweets in 2011 enhanced the platform's impact by offering a visual dimension to help amplify the reach of the messages. Images can serve as powerful tools to evoke emotional responses, clarify complex issues, and influence perceptions, but that is not always the case. Images can also be added just as an attention-grabbing strategy or clickbait, or even expressing humor as a meme. A tweet accompanied by a striking or controversial image can dramatically shift how readers interpret the message, adding layers of meaning or even altering the context. In this way, the synergy between text and visuals on the social network not only grabs attention but also guides the overall narrative.

In this work, we build upon previous research by \citet{vempala-preotiuc-pietro-2019-categorizing} and \citet{rikters-etal-2024-annotations} who introduced a four-class strategy for classifying image-text relations from Twitter data. We evaluate vision-language models on the Text-Image Relationship in Tweets\footnote{\url{https://github.com/danielpreotiuc/text-image-relationship/}} (TIRT) data set and the Latvian Twitter Eater Corpus\footnote{\url{https://github.com/Usprogis/Latvian-Twitter-Eater-Corpus/}} (LTEC).

Concretely, we consider the setting proposed by the latter authors who performed initial experiments with the LLaVA \cite{liu2023llava}, which we significantly extend in terms of model selection, robustness and evaluation scheme. One particular issue we tackle is the class imbalance of the data, further dividing their test set into a class-balanced evaluation set to lessen the overarching dominance of specific classes. We also employ a professional translator to manually translate the evaluation set from Latvian into English to minimize the potential errors that could be introduced by using automatic translations for the vision-language model (VLM) experiments. We experiment with five different open-source VLM checkpoints that are capable of running on consumer hardware.

Our results show that 1) larger newer models like LLaVA-NeXT 13B and Llama 3.2 11B are capable of outperforming the baseline and even smaller models like Phi 3.5 4B are reasonably competitive; 2) some models are not very sensitive to the input language (LLaVA-NeXT 7B, Llama 3.2 11B, Qwen2-VL 7B) while others perform far better when the input is in English (LLaVA-NeXT 13B, Phi 3.5 4B); 3) the results from different VLMs can be sensitive to the domain or the particular evaluation set used, since Llama 3.2 11B was overwhelmingly the highest performer on the LTEC data, but lowest on the TIRT data, while Qwen2-VL 7B scored lowest on LTEC, but was competitive on TIRT.


\section{Related Work}

\citet{vempala-preotiuc-pietro-2019-categorizing} introduced the categorization schema for the relations between Tweet text and attached images that we are using in our experiments. They distinguish four different categories: 1) the image adds to the text meaning and the text is represented in the image (further in the paper we will denote this using the emoji combination \painting \ \check \  \paper \ \check \ ); 2) the image adds to the text meaning and the text is not represented in the image (\painting \ \check \  \paper \ \cross \ ); 3) the image does not add to the text meaning and the text is represented in the image (\painting \ \cross \  \paper \ \check \ ); and 4) the image does not add to the text meaning and the text is not represented in the image (\painting \ \cross \  \paper \ \cross \ ). They also release the corpus of 4472 tweet-image pairs and their manually annotated relation categories (of which 2942 are still available at the time of writing this paper) and analyze the user demographic traits linked to each of the four image tweeting categories in depth. For simplification, these categories can be broken down into two yes/no questions, which makes it easier for prompting VLMs, however, the authors did not perform any such experiments.

\citet{rikters-etal-2024-annotations} apply the image-tweet categorization schema introduced by \citet{vempala-preotiuc-pietro-2019-categorizing} on the Latvian Twitter Eater Corpus (LTEC) by annotating 812 tweets written in Latvian about topics related to food and eating. They use this dataset to test the zero-shot classification abilities of the LLaVA model, concretely of their versions 1.3 and 1.5 in sizes of 7B and 13B parameters. These models are tested both in the original dataset of Latvian tweets, and in a version which is automatically translated English. They report that the best results using LLaVA 1.5 with 7B parameters, reaching a 20.69\% prediction accuracy when evaluated on the original Latvian texts, and increasing up to 27.83\% when evaluated on the automatic English translations. We consider this to be our direct baseline.

\citet{winata2024worldcuisinesmassivescalebenchmarkmultilingual} release a massively multilingual data set of food-related text-image pairs for visual question answering by identifying dish names and their origins in 30 languages. They evaluate these tasks using various VLMs in multiple sizes and release open-source code for experiment reproduction. Their results show that closed proprietary online API systems show overall superior performance, however, open-source models in the 70B-90B parameter range can still be quite competitive.

\section{Proposed Approach}

In this work, we commit to a more detailed evaluation of the image-text relation classification task for the available Twitter data. We aim to compare the performance of several recent VLMs that can be run on a reasonable desktop setup using a single NVIDIA RTX 3090 GPU with 24GB of VRAM. In our evaluation, we consider the following model versions and sizes -- Llama 3.2 Vision \cite{dubey2024llama3herdmodels} 11B, LLaVa-NeXT Vicuna  \cite{li2024llavanextinterleavetacklingmultiimagevideo} 7B and 13B, 
Qwen2-VL \cite{bai2023qwenvlversatilevisionlanguagemodel} 7B, Phi 3.5 Vision \cite{abdin2024phi3technicalreporthighly} 4B. We load all models from Hugging Face using the following identifiers - "microsoft/Phi-3.5-vision-instruct", "llava-hf/llava-v1.6-vicuna-7b-hf", "llava-hf/llava-v1.6-vicuna-13b-hf", "meta-llama/Llama-3.2-11B-Vision-Instruct", "Qwen/Qwen2-VL-7B-Instruct."

Our evaluation is based on the LTEC image-text relation test set in Latvian and manually translated English. The test set is reduced in size in favor of a more balanced class distribution, enabling a fair evaluation. In addition to the overall class, we also present a separate evaluation of the two individual questions prompted to the models - Q1) is the image adding to the text meaning; and Q2) is the text represented to the image. 


To further improve classification results, the two obvious directions to explore would be in-context learning \cite{zong2024vliclbenchdevildetails} by providing several examples of the image-text relation task at each inference step, or fine-tuning the model checkpoints on the image-text relation task. Both are currently out of scope in our case, as they require significantly more computation resources and a dedicated training data set. In addition, not all of our selected models are capable of processing several input images, which is a requirement for in-context learning to function.


\section{Data Preparation}
\label{sec:data}

We noticed several flaws in the previous work which evaluated the image-text relations using VLMs. Firstly, the data set composition was skewed strongly towards two out of four classes, as shown in Table \ref{tab:eval-distribution} - the image adding to the text meaning and text being represented in the image class with 48.28\% of the data and a further 36.45\% for the image not adding to the text meaning and text being represented in the image class, which together make up 84.73\% of the evaluation data. Furthermore, they did not report separate results on each of the individual components that define the task (Q1 and Q2), although these were obtained by separately prompting the VLMs. Finally, the evaluation which achieved the highest accuracy result was performed on automatically translated texts, which could be erroneous and therefore making way for the potential of creating further unnecessary errors in the classification task.

\subsection{Evaluation Set Balancing}

We extracted a part of the 812 tweet set into a separate evaluation set of 350 tweets to have a more even distribution among the four classes. The main objective was to reduce the dominance of the first and third classes. A comparison of the new distribution with the full original data set is shown in Table \ref{tab:eval-distribution}. The selection includes all available data for the two classes with the fewest examples (\painting \check \paper \cross \ and \painting \cross \paper \cross) and a random selection of 113 tweets for the other two classes (\painting \check \paper \check \ and \painting \cross \paper \check).

\begin{table}[t]
    \begin{tabular}{c|r|c|r}
        Class & Tweets & Percentage & Before \\ \hline
        \painting \ \check \  \paper \  \check \  & 113 & 32.29\% & 48.28\% \\
        \painting \ \check \  \paper \  \cross \  & 72  & 20.57\% & 8.87\% \\
        \painting \ \cross \  \paper \  \check \  & 113 & 32.29\% & 36.45\% \\
        \painting \ \cross \  \paper \  \cross \  & 52  & 14.86\% & 6.40\%  
    \end{tabular}
    \caption{Evaluation set class distribution. \protect\painting \  represents the image adding to the text meaning, \protect\paper \  -- the text being represented in the image, and \protect\check \  and \protect\cross \  -- true or false respectively.}
    \label{tab:eval-distribution}
\end{table}

\subsection{Manual Translation}

The highest text-image relation classification accuracy scores reported by \citet{rikters-etal-2024-annotations} were achieved by automatically translating the Latvian texts into English using an MT system that reaches scores of 48.28 BLEU and 68.21 ChrF on a separate evaluation set. While MT systems of such quality are generally usable, they are still far from perfect. To minimize the potential of error propagation, we employed a human translator to perform a full manual translation of the image-tweet relation texts from Latvian into English. We also evaluated three online systems\footnote{Tilde MT, Google Translate, DeepL Translate - all accessed in November 2024} and one open-source model\footnote{Opus MT tc-big-lv-en: \url{https://huggingface.co/Helsinki-NLP/opus-mt-tc-big-lv-en}} on the manually translated texts. Results in Table \ref{tab:mt-results} show that for this set, Google Translate outperforms all others in terms of BLEU \cite{papineni-etal-2002-bleu}, ChrF \cite{popovic-2015-chrf} and COMET \cite{rei-etal-2020-comet}, while Tilde MT, which was used in the evaluation of \citet{rikters-etal-2024-annotations}, scores the lowest. In the subsequent evaluations of this paper, we only use our manual translations of the Latvian tweets when referring to the English translations.

\begin{table}[t]
    \begin{tabular}{l|c|c|c}
              System & BLEU   & ChrF   & COMET  \\
                     \hline
    Tilde MT         & 52.63  & 67.94  & 78.50 \\
    Google Translate & \textbf{63.49}  & \textbf{75.56}  & \textbf{83.99} \\
    DeepL Translate  & 59.19  & 72.20  & 83.31 \\
    Opus MT          & 54.50  & 68.77  & 78.78         
    \end{tabular}
    \caption{Machine translation results. }
    \label{tab:mt-results}
\end{table}

\subsection{Instruction Formatting}

\begin{table*}[]
    \centering
    \begin{tabular}{c|c|l|l|l|l}
    \textbf{Prompt} & \textbf{Data} & \textbf{Model}     & \textbf{Class}        & \textbf{Question 1}           & \textbf{Question 2}            \\
    \hline
         &    & LLaVA-NeXT 7B  & 23.40 $\pm$ \ 8.03 & 51.57 $\pm$ \ 3.57 & 41.37 $\pm$ \ 21.49 \\
         &    & LLaVA-NeXT 13B & 19.43 $\pm$ \ 4.57 & 51.11 $\pm$ \ 6.03 & 34.60 $\pm$ \ 3.11 \\
  EN     & LV & Phi 3.5 4B    & 18.14 $\pm$ \ 3.00 & 48.49 $\pm$ \ 1.63 & 38.71 $\pm$ \ 3.57  \\
         &    & Qwen2-VL 7B   & 15.71 $\pm$ \ 0.00 & 47.71 $\pm$ \ 0.00 & 35.43 $\pm$ \ 0.00  \\
         &    & Llama 3.2 11B & \textbf{33.07} $\pm$ \ 0.36 & \underline{52.29} $\pm$ \ 0.29 & \textbf{69.21} $\pm$ \ 0.21   \\
    \hline
    \hline
    EN     & EN   & Baseline \citet{rikters-etal-2024-annotations} & 25.71 $\pm$ \ 4.00 & 52.77 $\pm$ \ 3.51 & 45.31 $\pm$ \ 4.11 \\
    \hline
    \hline
         &    & LLaVA-NeXT 7B  & 24.46 $\pm$ \ 7.83 & 52.17 $\pm$ \ 1.31 & 43.86 $\pm$ \ 18.71 \\
         &    & LLaVA-NeXT 13B & \underline{28.91} $\pm$ \ 6.34 & \textbf{53.20} $\pm$ \ 4.06 & \underline{51.40} $\pm$ \ 10.89 \\
  EN     & EN & Phi 3.5 4B    & 25.14 $\pm$ \ 5.71 & 48.31 $\pm$ \ 2.83 & 49.14 $\pm$ \ 7.43  \\
         &    & Qwen2-VL 7B   & 15.71 $\pm$ \ 0.00 & 47.43 $\pm$ \ 0.00 & 37.14 $\pm$ \ 0.00  \\
         &    & Llama 3.2 11B & \textbf{33.83} $\pm$ \ 0.17 & \underline{52.11} $\pm$ \ 0.17 & \textbf{66.77} $\pm$ \ 0.20 
    \end{tabular}
    \caption{Average classification accuracy results from zero-shot experiments using 10 different random seeds on the balanced subset of 350 selected Tweets from LTEC. Our baseline is the highest scoring run from \citet{rikters-etal-2024-annotations} using the LLaVA 1.5 model with 7B parameters. The highest results are marked in a \textbf{bold} font and the second highest are \underline{underlined}.}
    \label{tab:0-shot-result-table}
\end{table*}

It is well known that many modern large language models and therefore also VLMs can often be very sensitive to the provided prompt for a specific task and produce vastly variable results. In our experiments, we mainly kept using the prompt suggested by \citet{rikters-etal-2024-annotations} for all models except Llama 3.2, which required a very specific prompting approach to achieve consistent results. For that model we added the following text to the end of the prompt: Format the answer in the pattern of `**Answer:** {YES/NO}; **EXPLANATION:** {Motivation for the choosing the answer}'".

We also ran experiments with providing the instruction prompt in Latvian, however, for all models in large portions of the examples the generated answers were gibberish word salad, repetitions, empty strings or otherwise unquantifiable outputs as opposed to the expected ``YES/NO" answers. Therefore, we only report results using the instruction prompt in English and variations of tweet text language between Latvian and English.

\section{Results}
\label{sec:results}

\begin{table}[]
    \begin{tabular}{l|c|c|c}
             Model & Class   & Q1      & Q2      \\
                   \hline
    LLaVA-NeXT 7B  & 31.11 & 48.22 & \textbf{66.67} \\
    LLaVA-NeXT 13B & \textbf{39.11} & \underline{57.78} & \underline{65.11} \\
    Qwen2-VL       & 33.11 & 55.56 & 59.11 \\
    Phi 3.5        & \underline{36.44} & \textbf{63.78} & 57.56 \\
    Llama 3.2      & 22.22 & 44.44 & 46.00 \\
    \end{tabular}
    \caption{Evaluation results using a 450 tweet sample set from the TIRT data. The highest results are marked in a \textbf{bold} font and the second highest are \underline{underlined}.}
    \label{tab:alt-rez}
\end{table}

Our main results are summarized in Table \ref{tab:0-shot-result-table}. We compare five different models which represent 3 main size categories of 4B, 7B and 11B-13B parameters. Each evaluation is run 10 times with different seeds (the same 10 seeds for each model) with the prompt written in English and the actual tweet text provided in either Latvian or English. We compare classification accuracy on the overall class, as well as each of the two individual questions of the image adding to the meaning and text being represented in the image. 

The result table shows a large variation in both the overall class accuracy, and in the individual questions. Llama 3.2 is clearly the highest performer regardless of the language of the input text, followed by the LLaVA-NeXT models and Phi 3.5, of which all seem to prefer the English translation rather than the original Latvian text. Qwen2-VL scores the lowest, regardless of the input language, and also exhibits no variation with the different random seeds. Meanwhile, Llama 3.2 shows only a very small sensitivity to random seed changes, but Phi 3.5 and especially LLaVA-NeXT models tend to vary a lot. Both Llama 3.2 11B and LLaVA-NeXT 13B outperform the baseline results, however only the result from Llama 3.2 11B is statistically significant.


For comparison, we also sampled a random subset of 450 tweets from the larger TIRT data set for evaluation. This data set seems to be naturally much better distributed, having a class distribution of 19.33\% : 24.89\% : 23.33\% : 32.45\%. Classification accuracy results in Table \ref{tab:alt-rez} show overall higher scores than the domain-specific Latvian food tweet LTEC data set. However, the results are still relatively low and have the potential to be further improved. Interestingly, Llama 3.2 11B was the worst overall performer on this set and Qwen2-VL 7B, which was the worst on LTEC, fared much better on TIRT. 

The results from both tables demonstrate the overall robustness of the LLaVA-NeXT 13B and Phi 3.5 4B models, as long as the input text is provided in English.






\section{Conclusion}

In this paper, we introduced an extended evaluation of the image-text relation task for social media posts from Twitter. We prepared a balanced version of a previously available image-text relation data set, as well as a manual English translation of its original texts in the Latvian language. We experimented with various open-source vision-language models and demonstrated how results vary depending on multiple conditions. Our findings show that LLaVA-NeXT 13B and Phi 3.5 4B models can handle this task on both evaluation sets very well as long as the input text is provided in English. Meanwhile Llama 3.2 11B and Qwen2-VL 7B are more robust towards input language, but very sensitive to the input data domain.

We plan to release our balanced evaluation data set along with evaluation code for easy reproduction of our results or similar experiments. In future work we plan to perform experiments using in-context learning and model fine-tuning on the image-text relation task.

\section*{Limitations}

In this work, we only considered using data and models that are publicly available for research purposes to enable reproducibility. Also, since running 70+ billion parameter sized large models is computationally very costly, we opt for choosing models with fewer parameters in our experiments.

\section*{Ethical Considerations}

Our work is fully in accordance with the ACL Code of Ethics\footnote{\url{https://www.aclweb.org/portal/content/acl-code-ethics}}. We use only publicly available datasets and relatively low compute amounts while conducting our experiments to enable reproducibility. We do not conduct studies on other humans or animals in this research.



\bibliographystyle{acl_natbib}
\bibliography{anthology,custom}

\begin{thebibliography}{12}
\expandafter\ifx\csname natexlab\endcsname\relax\def\natexlab#1{#1}\fi

\bibitem[{Abdin et~al.(2024)Abdin, Aneja, Awadalla, Awadallah, Awan, Bach, Bahree, Bakhtiari, Bao, Behl, Benhaim, Bilenko, Bjorck, Bubeck, Cai, Cai, Chaudhary, Chen, Chen, Chen, Chen, Chen, Cheng, Chopra, Dai, Dixon, Eldan, Fragoso, Gao, Gao, Gao, Garg, Giorno, Goswami, Gunasekar, Haider, Hao, Hewett, Hu, Huynh, Iter, Jacobs, Javaheripi, Jin, Karampatziakis, Kauffmann, Khademi, Kim, Kim, Kurilenko, Lee, Lee, Li, Li, Liang, Liden, Lin, Lin, Liu, Liu, Liu, Liu, Liu, Luo, Madan, Mahmoudzadeh, Majercak, Mazzola, Mendes, Mitra, Modi, Nguyen, Norick, Patra, Perez-Becker, Portet, Pryzant, Qin, Radmilac, Ren, de~Rosa, Rosset, Roy, Ruwase, Saarikivi, Saied, Salim, Santacroce, Shah, Shang, Sharma, Shen, Shukla, Song, Tanaka, Tupini, Vaddamanu, Wang, Wang, Wang, Wang, Wang, Wang, Ward, Wen, Witte, Wu, Wu, Wyatt, Xiao, Xu, Xu, Xu, Xue, Yadav, Yang, Yang, Yang, Yang, Yu, Yuan, Zhang, Zhang, Zhang, Zhang, Zhang, Zhang, Zhang, and Zhou}]{abdin2024phi3technicalreporthighly}
Marah Abdin, Jyoti Aneja, Hany Awadalla, Ahmed Awadallah, Ammar~Ahmad Awan, Nguyen Bach, Amit Bahree, Arash Bakhtiari, Jianmin Bao, Harkirat Behl, Alon Benhaim, Misha Bilenko, Johan Bjorck, Sébastien Bubeck, Martin Cai, Qin Cai, Vishrav Chaudhary, Dong Chen, Dongdong Chen, Weizhu Chen, Yen-Chun Chen, Yi-Ling Chen, Hao Cheng, Parul Chopra, Xiyang Dai, Matthew Dixon, Ronen Eldan, Victor Fragoso, Jianfeng Gao, Mei Gao, Min Gao, Amit Garg, Allie~Del Giorno, Abhishek Goswami, Suriya Gunasekar, Emman Haider, Junheng Hao, Russell~J. Hewett, Wenxiang Hu, Jamie Huynh, Dan Iter, Sam~Ade Jacobs, Mojan Javaheripi, Xin Jin, Nikos Karampatziakis, Piero Kauffmann, Mahoud Khademi, Dongwoo Kim, Young~Jin Kim, Lev Kurilenko, James~R. Lee, Yin~Tat Lee, Yuanzhi Li, Yunsheng Li, Chen Liang, Lars Liden, Xihui Lin, Zeqi Lin, Ce~Liu, Liyuan Liu, Mengchen Liu, Weishung Liu, Xiaodong Liu, Chong Luo, Piyush Madan, Ali Mahmoudzadeh, David Majercak, Matt Mazzola, Caio César~Teodoro Mendes, Arindam Mitra, Hardik Modi, Anh Nguyen,
  Brandon Norick, Barun Patra, Daniel Perez-Becker, Thomas Portet, Reid Pryzant, Heyang Qin, Marko Radmilac, Liliang Ren, Gustavo de~Rosa, Corby Rosset, Sambudha Roy, Olatunji Ruwase, Olli Saarikivi, Amin Saied, Adil Salim, Michael Santacroce, Shital Shah, Ning Shang, Hiteshi Sharma, Yelong Shen, Swadheen Shukla, Xia Song, Masahiro Tanaka, Andrea Tupini, Praneetha Vaddamanu, Chunyu Wang, Guanhua Wang, Lijuan Wang, Shuohang Wang, Xin Wang, Yu~Wang, Rachel Ward, Wen Wen, Philipp Witte, Haiping Wu, Xiaoxia Wu, Michael Wyatt, Bin Xiao, Can Xu, Jiahang Xu, Weijian Xu, Jilong Xue, Sonali Yadav, Fan Yang, Jianwei Yang, Yifan Yang, Ziyi Yang, Donghan Yu, Lu~Yuan, Chenruidong Zhang, Cyril Zhang, Jianwen Zhang, Li~Lyna Zhang, Yi~Zhang, Yue Zhang, Yunan Zhang, and Xiren Zhou. 2024.
\newblock \href {http://arxiv.org/abs/2404.14219} {Phi-3 technical report: A highly capable language model locally on your phone}.

\bibitem[{Bai et~al.(2023)Bai, Bai, Yang, Wang, Tan, Wang, Lin, Zhou, and Zhou}]{bai2023qwenvlversatilevisionlanguagemodel}
Jinze Bai, Shuai Bai, Shusheng Yang, Shijie Wang, Sinan Tan, Peng Wang, Junyang Lin, Chang Zhou, and Jingren Zhou. 2023.
\newblock \href {http://arxiv.org/abs/2308.12966} {Qwen-vl: A versatile vision-language model for understanding, localization, text reading, and beyond}.

\bibitem[{Dubey et~al.(2024)Dubey, Jauhri, Pandey, Kadian, Al-Dahle, Letman, Mathur, Schelten, Yang, Fan, Goyal, Hartshorn, Yang, Mitra, Sravankumar, Korenev, Hinsvark, Rao, Zhang, Rodriguez, Gregerson, Spataru, Roziere, Biron, Tang, Chern, Caucheteux, Nayak, Bi, Marra, McConnell, Keller, Touret, Wu, Wong, Ferrer, Nikolaidis, Allonsius, Song, Pintz, Livshits, Esiobu, Choudhary, Mahajan, Garcia-Olano, Perino, Hupkes, Lakomkin, AlBadawy, Lobanova, Dinan, Smith, Radenovic, Zhang, Synnaeve, Lee, Anderson, Nail, Mialon, Pang, Cucurell, Nguyen, Korevaar, Xu, Touvron, Zarov, Ibarra, Kloumann, Misra, Evtimov, Copet, Lee, Geffert, Vranes, Park, Mahadeokar, Shah, van~der Linde, Billock, Hong, Lee, Fu, Chi, Huang, Liu, Wang, Yu, Bitton, Spisak, Park, Rocca, Johnstun, Saxe, Jia, Alwala, Upasani, Plawiak, Li, Heafield, Stone, El-Arini, Iyer, Malik, Chiu, Bhalla, Rantala-Yeary, van~der Maaten, Chen, Tan, Jenkins, Martin, Madaan, Malo, Blecher, Landzaat, de~Oliveira, Muzzi, Pasupuleti, Singh, Paluri, Kardas, Oldham, Rita,
  Pavlova, Kambadur, Lewis, Si, Singh, Hassan, Goyal, Torabi, Bashlykov, Bogoychev, Chatterji, Duchenne, Çelebi, Alrassy, Zhang, Li, Vasic, Weng, Bhargava, Dubal, Krishnan, Koura, Xu, He, Dong, Srinivasan, Ganapathy, Calderer, Cabral, Stojnic, Raileanu, Girdhar, Patel, Sauvestre, Polidoro, Sumbaly, Taylor, Silva, Hou, Wang, Hosseini, Chennabasappa, Singh, Bell, Kim, Edunov, Nie, Narang, Raparthy, Shen, Wan, Bhosale, Zhang, Vandenhende, Batra, Whitman, Sootla, Collot, Gururangan, Borodinsky, Herman, Fowler, Sheasha, Georgiou, Scialom, Speckbacher, Mihaylov, Xiao, Karn, Goswami, Gupta, Ramanathan, Kerkez, Gonguet, Do, Vogeti, Petrovic, Chu, Xiong, Fu, Meers, Martinet, Wang, Tan, Xie, Jia, Wang, Goldschlag, Gaur, Babaei, Wen, Song, Zhang, Li, Mao, Coudert, Yan, Chen, Papakipos, Singh, Grattafiori, Jain, Kelsey, Shajnfeld, Gangidi, Victoria, Goldstand, Menon, Sharma, Boesenberg, Vaughan, Baevski, Feinstein, Kallet, Sangani, Yunus, Lupu, Alvarado, Caples, Gu, Ho, Poulton, Ryan, Ramchandani, Franco, Saraf,
  Chowdhury, Gabriel, Bharambe, Eisenman, Yazdan, James, Maurer, Leonhardi, Huang, Loyd, Paola, Paranjape, Liu, Wu, Ni, Hancock, Wasti, Spence, Stojkovic, Gamido, Montalvo, Parker, Burton, Mejia, Wang, Kim, Zhou, Hu, Chu, Cai, Tindal, Feichtenhofer, Civin, Beaty, Kreymer, Li, Wyatt, Adkins, Xu, Testuggine, David, Parikh, Liskovich, Foss, Wang, Le, Holland, Dowling, Jamil, Montgomery, Presani, Hahn, Wood, Brinkman, Arcaute, Dunbar, Smothers, Sun, Kreuk, Tian, Ozgenel, Caggioni, Guzmán, Kanayet, Seide, Florez, Schwarz, Badeer, Swee, Halpern, Thattai, Herman, Sizov, Guangyi, Zhang, Lakshminarayanan, Shojanazeri, Zou, Wang, Zha, Habeeb, Rudolph, Suk, Aspegren, Goldman, Damlaj, Molybog, Tufanov, Veliche, Gat, Weissman, Geboski, Kohli, Asher, Gaya, Marcus, Tang, Chan, Zhen, Reizenstein, Teboul, Zhong, Jin, Yang, Cummings, Carvill, Shepard, McPhie, Torres, Ginsburg, Wang, Wu, U, Saxena, Prasad, Khandelwal, Zand, Matosich, Veeraraghavan, Michelena, Li, Huang, Chawla, Lakhotia, Huang, Chen, Garg, A, Silva, Bell,
  Zhang, Guo, Yu, Moshkovich, Wehrstedt, Khabsa, Avalani, Bhatt, Tsimpoukelli, Mankus, Hasson, Lennie, Reso, Groshev, Naumov, Lathi, Keneally, Seltzer, Valko, Restrepo, Patel, Vyatskov, Samvelyan, Clark, Macey, Wang, Hermoso, Metanat, Rastegari, Bansal, Santhanam, Parks, White, Bawa, Singhal, Egebo, Usunier, Laptev, Dong, Zhang, Cheng, Chernoguz, Hart, Salpekar, Kalinli, Kent, Parekh, Saab, Balaji, Rittner, Bontrager, Roux, Dollar, Zvyagina, Ratanchandani, Yuvraj, Liang, Alao, Rodriguez, Ayub, Murthy, Nayani, Mitra, Li, Hogan, Battey, Wang, Maheswari, Howes, Rinott, Bondu, Datta, Chugh, Hunt, Dhillon, Sidorov, Pan, Verma, Yamamoto, Ramaswamy, Lindsay, Lindsay, Feng, Lin, Zha, Shankar, Zhang, Zhang, Wang, Agarwal, Sajuyigbe, Chintala, Max, Chen, Kehoe, Satterfield, Govindaprasad, Gupta, Cho, Virk, Subramanian, Choudhury, Goldman, Remez, Glaser, Best, Kohler, Robinson, Li, Zhang, Matthews, Chou, Shaked, Vontimitta, Ajayi, Montanez, Mohan, Kumar, Mangla, Albiero, Ionescu, Poenaru, Mihailescu, Ivanov, Li, Wang,
  Jiang, Bouaziz, Constable, Tang, Wang, Wu, Wang, Xia, Wu, Gao, Chen, Hu, Jia, Qi, Li, Zhang, Zhang, Adi, Nam, Yu, Wang, Hao, Qian, He, Rait, DeVito, Rosnbrick, Wen, Yang, and Zhao}]{dubey2024llama3herdmodels}
Abhimanyu Dubey, Abhinav Jauhri, Abhinav Pandey, Abhishek Kadian, Ahmad Al-Dahle, Aiesha Letman, Akhil Mathur, Alan Schelten, Amy Yang, Angela Fan, Anirudh Goyal, Anthony Hartshorn, Aobo Yang, Archi Mitra, Archie Sravankumar, Artem Korenev, Arthur Hinsvark, Arun Rao, Aston Zhang, Aurelien Rodriguez, Austen Gregerson, Ava Spataru, Baptiste Roziere, Bethany Biron, Binh Tang, Bobbie Chern, Charlotte Caucheteux, Chaya Nayak, Chloe Bi, Chris Marra, Chris McConnell, Christian Keller, Christophe Touret, Chunyang Wu, Corinne Wong, Cristian~Canton Ferrer, Cyrus Nikolaidis, Damien Allonsius, Daniel Song, Danielle Pintz, Danny Livshits, David Esiobu, Dhruv Choudhary, Dhruv Mahajan, Diego Garcia-Olano, Diego Perino, Dieuwke Hupkes, Egor Lakomkin, Ehab AlBadawy, Elina Lobanova, Emily Dinan, Eric~Michael Smith, Filip Radenovic, Frank Zhang, Gabriel Synnaeve, Gabrielle Lee, Georgia~Lewis Anderson, Graeme Nail, Gregoire Mialon, Guan Pang, Guillem Cucurell, Hailey Nguyen, Hannah Korevaar, Hu~Xu, Hugo Touvron, Iliyan Zarov,
  Imanol~Arrieta Ibarra, Isabel Kloumann, Ishan Misra, Ivan Evtimov, Jade Copet, Jaewon Lee, Jan Geffert, Jana Vranes, Jason Park, Jay Mahadeokar, Jeet Shah, Jelmer van~der Linde, Jennifer Billock, Jenny Hong, Jenya Lee, Jeremy Fu, Jianfeng Chi, Jianyu Huang, Jiawen Liu, Jie Wang, Jiecao Yu, Joanna Bitton, Joe Spisak, Jongsoo Park, Joseph Rocca, Joshua Johnstun, Joshua Saxe, Junteng Jia, Kalyan~Vasuden Alwala, Kartikeya Upasani, Kate Plawiak, Ke~Li, Kenneth Heafield, Kevin Stone, Khalid El-Arini, Krithika Iyer, Kshitiz Malik, Kuenley Chiu, Kunal Bhalla, Lauren Rantala-Yeary, Laurens van~der Maaten, Lawrence Chen, Liang Tan, Liz Jenkins, Louis Martin, Lovish Madaan, Lubo Malo, Lukas Blecher, Lukas Landzaat, Luke de~Oliveira, Madeline Muzzi, Mahesh Pasupuleti, Mannat Singh, Manohar Paluri, Marcin Kardas, Mathew Oldham, Mathieu Rita, Maya Pavlova, Melanie Kambadur, Mike Lewis, Min Si, Mitesh~Kumar Singh, Mona Hassan, Naman Goyal, Narjes Torabi, Nikolay Bashlykov, Nikolay Bogoychev, Niladri Chatterji, Olivier
  Duchenne, Onur Çelebi, Patrick Alrassy, Pengchuan Zhang, Pengwei Li, Petar Vasic, Peter Weng, Prajjwal Bhargava, Pratik Dubal, Praveen Krishnan, Punit~Singh Koura, Puxin Xu, Qing He, Qingxiao Dong, Ragavan Srinivasan, Raj Ganapathy, Ramon Calderer, Ricardo~Silveira Cabral, Robert Stojnic, Roberta Raileanu, Rohit Girdhar, Rohit Patel, Romain Sauvestre, Ronnie Polidoro, Roshan Sumbaly, Ross Taylor, Ruan Silva, Rui Hou, Rui Wang, Saghar Hosseini, Sahana Chennabasappa, Sanjay Singh, Sean Bell, Seohyun~Sonia Kim, Sergey Edunov, Shaoliang Nie, Sharan Narang, Sharath Raparthy, Sheng Shen, Shengye Wan, Shruti Bhosale, Shun Zhang, Simon Vandenhende, Soumya Batra, Spencer Whitman, Sten Sootla, Stephane Collot, Suchin Gururangan, Sydney Borodinsky, Tamar Herman, Tara Fowler, Tarek Sheasha, Thomas Georgiou, Thomas Scialom, Tobias Speckbacher, Todor Mihaylov, Tong Xiao, Ujjwal Karn, Vedanuj Goswami, Vibhor Gupta, Vignesh Ramanathan, Viktor Kerkez, Vincent Gonguet, Virginie Do, Vish Vogeti, Vladan Petrovic, Weiwei Chu,
  Wenhan Xiong, Wenyin Fu, Whitney Meers, Xavier Martinet, Xiaodong Wang, Xiaoqing~Ellen Tan, Xinfeng Xie, Xuchao Jia, Xuewei Wang, Yaelle Goldschlag, Yashesh Gaur, Yasmine Babaei, Yi~Wen, Yiwen Song, Yuchen Zhang, Yue Li, Yuning Mao, Zacharie~Delpierre Coudert, Zheng Yan, Zhengxing Chen, Zoe Papakipos, Aaditya Singh, Aaron Grattafiori, Abha Jain, Adam Kelsey, Adam Shajnfeld, Adithya Gangidi, Adolfo Victoria, Ahuva Goldstand, Ajay Menon, Ajay Sharma, Alex Boesenberg, Alex Vaughan, Alexei Baevski, Allie Feinstein, Amanda Kallet, Amit Sangani, Anam Yunus, Andrei Lupu, Andres Alvarado, Andrew Caples, Andrew Gu, Andrew Ho, Andrew Poulton, Andrew Ryan, Ankit Ramchandani, Annie Franco, Aparajita Saraf, Arkabandhu Chowdhury, Ashley Gabriel, Ashwin Bharambe, Assaf Eisenman, Azadeh Yazdan, Beau James, Ben Maurer, Benjamin Leonhardi, Bernie Huang, Beth Loyd, Beto~De Paola, Bhargavi Paranjape, Bing Liu, Bo~Wu, Boyu Ni, Braden Hancock, Bram Wasti, Brandon Spence, Brani Stojkovic, Brian Gamido, Britt Montalvo, Carl
  Parker, Carly Burton, Catalina Mejia, Changhan Wang, Changkyu Kim, Chao Zhou, Chester Hu, Ching-Hsiang Chu, Chris Cai, Chris Tindal, Christoph Feichtenhofer, Damon Civin, Dana Beaty, Daniel Kreymer, Daniel Li, Danny Wyatt, David Adkins, David Xu, Davide Testuggine, Delia David, Devi Parikh, Diana Liskovich, Didem Foss, Dingkang Wang, Duc Le, Dustin Holland, Edward Dowling, Eissa Jamil, Elaine Montgomery, Eleonora Presani, Emily Hahn, Emily Wood, Erik Brinkman, Esteban Arcaute, Evan Dunbar, Evan Smothers, Fei Sun, Felix Kreuk, Feng Tian, Firat Ozgenel, Francesco Caggioni, Francisco Guzmán, Frank Kanayet, Frank Seide, Gabriela~Medina Florez, Gabriella Schwarz, Gada Badeer, Georgia Swee, Gil Halpern, Govind Thattai, Grant Herman, Grigory Sizov, Guangyi, Zhang, Guna Lakshminarayanan, Hamid Shojanazeri, Han Zou, Hannah Wang, Hanwen Zha, Haroun Habeeb, Harrison Rudolph, Helen Suk, Henry Aspegren, Hunter Goldman, Ibrahim Damlaj, Igor Molybog, Igor Tufanov, Irina-Elena Veliche, Itai Gat, Jake Weissman, James
  Geboski, James Kohli, Japhet Asher, Jean-Baptiste Gaya, Jeff Marcus, Jeff Tang, Jennifer Chan, Jenny Zhen, Jeremy Reizenstein, Jeremy Teboul, Jessica Zhong, Jian Jin, Jingyi Yang, Joe Cummings, Jon Carvill, Jon Shepard, Jonathan McPhie, Jonathan Torres, Josh Ginsburg, Junjie Wang, Kai Wu, Kam~Hou U, Karan Saxena, Karthik Prasad, Kartikay Khandelwal, Katayoun Zand, Kathy Matosich, Kaushik Veeraraghavan, Kelly Michelena, Keqian Li, Kun Huang, Kunal Chawla, Kushal Lakhotia, Kyle Huang, Lailin Chen, Lakshya Garg, Lavender A, Leandro Silva, Lee Bell, Lei Zhang, Liangpeng Guo, Licheng Yu, Liron Moshkovich, Luca Wehrstedt, Madian Khabsa, Manav Avalani, Manish Bhatt, Maria Tsimpoukelli, Martynas Mankus, Matan Hasson, Matthew Lennie, Matthias Reso, Maxim Groshev, Maxim Naumov, Maya Lathi, Meghan Keneally, Michael~L. Seltzer, Michal Valko, Michelle Restrepo, Mihir Patel, Mik Vyatskov, Mikayel Samvelyan, Mike Clark, Mike Macey, Mike Wang, Miquel~Jubert Hermoso, Mo~Metanat, Mohammad Rastegari, Munish Bansal, Nandhini
  Santhanam, Natascha Parks, Natasha White, Navyata Bawa, Nayan Singhal, Nick Egebo, Nicolas Usunier, Nikolay~Pavlovich Laptev, Ning Dong, Ning Zhang, Norman Cheng, Oleg Chernoguz, Olivia Hart, Omkar Salpekar, Ozlem Kalinli, Parkin Kent, Parth Parekh, Paul Saab, Pavan Balaji, Pedro Rittner, Philip Bontrager, Pierre Roux, Piotr Dollar, Polina Zvyagina, Prashant Ratanchandani, Pritish Yuvraj, Qian Liang, Rachad Alao, Rachel Rodriguez, Rafi Ayub, Raghotham Murthy, Raghu Nayani, Rahul Mitra, Raymond Li, Rebekkah Hogan, Robin Battey, Rocky Wang, Rohan Maheswari, Russ Howes, Ruty Rinott, Sai~Jayesh Bondu, Samyak Datta, Sara Chugh, Sara Hunt, Sargun Dhillon, Sasha Sidorov, Satadru Pan, Saurabh Verma, Seiji Yamamoto, Sharadh Ramaswamy, Shaun Lindsay, Shaun Lindsay, Sheng Feng, Shenghao Lin, Shengxin~Cindy Zha, Shiva Shankar, Shuqiang Zhang, Shuqiang Zhang, Sinong Wang, Sneha Agarwal, Soji Sajuyigbe, Soumith Chintala, Stephanie Max, Stephen Chen, Steve Kehoe, Steve Satterfield, Sudarshan Govindaprasad, Sumit Gupta,
  Sungmin Cho, Sunny Virk, Suraj Subramanian, Sy~Choudhury, Sydney Goldman, Tal Remez, Tamar Glaser, Tamara Best, Thilo Kohler, Thomas Robinson, Tianhe Li, Tianjun Zhang, Tim Matthews, Timothy Chou, Tzook Shaked, Varun Vontimitta, Victoria Ajayi, Victoria Montanez, Vijai Mohan, Vinay~Satish Kumar, Vishal Mangla, Vítor Albiero, Vlad Ionescu, Vlad Poenaru, Vlad~Tiberiu Mihailescu, Vladimir Ivanov, Wei Li, Wenchen Wang, Wenwen Jiang, Wes Bouaziz, Will Constable, Xiaocheng Tang, Xiaofang Wang, Xiaojian Wu, Xiaolan Wang, Xide Xia, Xilun Wu, Xinbo Gao, Yanjun Chen, Ye~Hu, Ye~Jia, Ye~Qi, Yenda Li, Yilin Zhang, Ying Zhang, Yossi Adi, Youngjin Nam, Yu, Wang, Yuchen Hao, Yundi Qian, Yuzi He, Zach Rait, Zachary DeVito, Zef Rosnbrick, Zhaoduo Wen, Zhenyu Yang, and Zhiwei Zhao. 2024.
\newblock \href {http://arxiv.org/abs/2407.21783} {The llama 3 herd of models}.

\bibitem[{Li et~al.(2024)Li, Zhang, Zhang, Zhang, Li, Li, Ma, and Li}]{li2024llavanextinterleavetacklingmultiimagevideo}
Feng Li, Renrui Zhang, Hao Zhang, Yuanhan Zhang, Bo~Li, Wei Li, Zejun Ma, and Chunyuan Li. 2024.
\newblock \href {http://arxiv.org/abs/2407.07895} {Llava-next-interleave: Tackling multi-image, video, and 3d in large multimodal models}.

\bibitem[{Liu et~al.(2023)Liu, Li, Wu, and Lee}]{liu2023llava}
Haotian Liu, Chunyuan Li, Qingyang Wu, and Yong~Jae Lee. 2023.
\newblock \href {https://openreview.net/forum?id=w0H2xGHlkw} {Visual instruction tuning}.
\newblock In \emph{Thirty-seventh Conference on Neural Information Processing Systems}.

\bibitem[{Papineni et~al.(2002)Papineni, Roukos, Ward, and Zhu}]{papineni-etal-2002-bleu}
Kishore Papineni, Salim Roukos, Todd Ward, and Wei-Jing Zhu. 2002.
\newblock \href {https://doi.org/10.3115/1073083.1073135} {{B}leu: a method for automatic evaluation of machine translation}.
\newblock In \emph{Proceedings of the 40th Annual Meeting of the Association for Computational Linguistics}, pages 311--318, Philadelphia, Pennsylvania, USA. Association for Computational Linguistics.

\bibitem[{Popovi{\'c}(2015)}]{popovic-2015-chrf}
Maja Popovi{\'c}. 2015.
\newblock \href {https://doi.org/10.18653/v1/W15-3049} {chr{F}: character n-gram {F}-score for automatic {MT} evaluation}.
\newblock In \emph{Proceedings of the Tenth Workshop on Statistical Machine Translation}, pages 392--395, Lisbon, Portugal. Association for Computational Linguistics.

\bibitem[{Rei et~al.(2020)Rei, Stewart, Farinha, and Lavie}]{rei-etal-2020-comet}
Ricardo Rei, Craig Stewart, Ana~C Farinha, and Alon Lavie. 2020.
\newblock \href {https://doi.org/10.18653/v1/2020.emnlp-main.213} {{COMET}: A neural framework for {MT} evaluation}.
\newblock In \emph{Proceedings of the 2020 Conference on Empirical Methods in Natural Language Processing (EMNLP)}, pages 2685--2702, Online. Association for Computational Linguistics.

\bibitem[{Rikters et~al.(2024)Rikters, V{\=\i}ksna, and Marrese-Taylor}]{rikters-etal-2024-annotations}
Mat\={\i}ss Rikters, Rinalds V{\=\i}ksna, and Edison Marrese-Taylor. 2024.
\newblock \href {https://aclanthology.org/2024.lrec-main.111} {Annotations for exploring food tweets from multiple aspects}.
\newblock In \emph{Proceedings of the 2024 Joint International Conference on Computational Linguistics, Language Resources and Evaluation (LREC-COLING 2024)}, pages 1233--1238, Torino, Italia. ELRA and ICCL.

\bibitem[{Vempala and Preo{\c{t}}iuc-Pietro(2019)}]{vempala-preotiuc-pietro-2019-categorizing}
Alakananda Vempala and Daniel Preo{\c{t}}iuc-Pietro. 2019.
\newblock \href {https://doi.org/10.18653/v1/P19-1272} {Categorizing and inferring the relationship between the text and image of {T}witter posts}.
\newblock In \emph{Proceedings of the 57th Annual Meeting of the Association for Computational Linguistics}, pages 2830--2840, Florence, Italy. Association for Computational Linguistics.

\bibitem[{Winata et~al.(2025)Winata, Hudi, Irawan, Anugraha, Putri, Yutong, Nohejl, Prathama, Ousidhoum, Amriani, Rzayev, Das, Pramodya, Adila, Wilie, Mawalim, Lam, Abolade, Chersoni, Santus, Ikhwantri, Kuwanto, Zhao, Wibowo, Lovenia, Cruz, Putra, Myung, Susanto, Machin, Zhukova, Anugraha, Adilazuarda, Santosa, Limkonchotiwat, Dabre, Audino, Cahyawijaya, Zhang, Salim, Zhou, Gui, Adelani, Lee, Okada, Purwarianti, Aji, Watanabe, Wijaya, Oh, and Ngo}]{winata2024worldcuisinesmassivescalebenchmarkmultilingual}
Genta~Indra Winata, Frederikus Hudi, Patrick~Amadeus Irawan, David Anugraha, Rifki~Afina Putri, Wang Yutong, Adam Nohejl, Ubaidillah~Ariq Prathama, Nedjma Ousidhoum, Afifa Amriani, Anar Rzayev, Anirban Das, Ashmari Pramodya, Aulia Adila, Bryan Wilie, Candy~Olivia Mawalim, Cheng~Ching Lam, Daud Abolade, Emmanuele Chersoni, Enrico Santus, Fariz Ikhwantri, Garry Kuwanto, Hanyang Zhao, Haryo~Akbarianto Wibowo, Holy Lovenia, Jan Christian~Blaise Cruz, Jan Wira~Gotama Putra, Junho Myung, Lucky Susanto, Maria Angelica~Riera Machin, Marina Zhukova, Michael Anugraha, Muhammad~Farid Adilazuarda, Natasha~Christabelle Santosa, Peerat Limkonchotiwat, Raj Dabre, Rio~Alexander Audino, Samuel Cahyawijaya, Shi-Xiong Zhang, Stephanie~Yulia Salim, Yi~Zhou, Yinxuan Gui, David~Ifeoluwa Adelani, En-Shiun~Annie Lee, Shogo Okada, Ayu Purwarianti, Alham~Fikri Aji, Taro Watanabe, Derry~Tanti Wijaya, Alice Oh, and Chong-Wah Ngo. 2025.
\newblock \href {https://aclanthology.org/2025.naacl-long.167/} {{W}orld{C}uisines: A massive-scale benchmark for multilingual and multicultural visual question answering on global cuisines}.
\newblock In \emph{Proceedings of the 2025 Conference of the Nations of the Americas Chapter of the Association for Computational Linguistics: Human Language Technologies (Volume 1: Long Papers)}, pages 3242--3264, Albuquerque, New Mexico. Association for Computational Linguistics.

\bibitem[{Zong et~al.(2024)Zong, Bohdal, and Hospedales}]{zong2024vliclbenchdevildetails}
Yongshuo Zong, Ondrej Bohdal, and Timothy Hospedales. 2024.
\newblock \href {http://arxiv.org/abs/2403.13164} {Vl-icl bench: The devil in the details of multimodal in-context learning}.

\end{thebibliography}

\appendix

\end{document}